\documentclass[sigconf]{acmart}
\usepackage[utf8]{inputenc}
\usepackage[T1]{fontenc}
\usepackage{graphicx}
\usepackage{caption}
\usepackage{cleveref}
\usepackage{booktabs}
\usepackage{tabularx}
\usepackage{tcolorbox}
\usepackage{enumitem}
\usepackage{hyperref}
\usepackage[table]{xcolor}

\AtBeginDocument{%
  }

\copyrightyear{2026}
\acmYear{2026}
\setcopyright{cc}
\setcctype{by}
\acmConference[WWW Companion '26]{Companion Proceedings of the ACM Web Conference 2026}{April 13--17, 2026}{Dubai, United Arab Emirates}
\acmBooktitle{Companion Proceedings of the ACM Web Conference 2026 (WWW Companion '26), April 13--17, 2026, Dubai, United Arab Emirates}
\acmPrice{}
\acmDOI{10.1145/3774905.3795449}
\acmISBN{979-8-4007-2308-7/2026/04}

\usepackage{microtype} 

\begin{document}

\title{Benchmarking Temporal Web3 Intelligence: Lessons from the FinSurvival 2025 Challenge}

\author{Oshani Seneviratne}
\email{senevo@rpi.edu}
\orcid{0000-0001-8518-917X}
\affiliation{%
  \institution{Rensselaer Polytechnic Institute}
  \city{Troy} \state{NY} \country{USA}
}

\author{Fernando Spadea}
\email{spadef@rpi.edu}
\orcid{0009-0006-4278-3666}
\affiliation{%
  \institution{Rensselaer Polytechnic Institute}
  \city{Troy} \state{NY} \country{USA}
}

\author{Adrien Pavao}
\email{pavao@mlchallenges.com}
\affiliation{%
  \institution{Codabench, MLChallenges}
  \city{Paris} \country{France}
}

\author{Aaron Micah Green}
\email{aarongreen@vassar.edu}
\affiliation{%
  \institution{Vassar College}
  \city{Poughkeepsie} \state{NY} \country{USA}
}

\author{Kristin P. Bennett}
\email{bennek@rpi.edu}
\affiliation{%
  \institution{Rensselaer Polytechnic Institute}
  \city{Troy} \state{NY} \country{USA}
}

\renewcommand{\shortauthors}{Oshani Seneviratne, Fernando Spadea, Adrien Pavao, Aaron Micah Green, \& Kristin P. Bennett}

\renewcommand{\shorttitle}{Lessons from the FinSurvival 2025 Challenge}

\begin{abstract}
Temporal Web analytics increasingly relies on large-scale, longitudinal data to understand how users, content, and systems evolve over time. A rapidly growing frontier is the \emph{Temporal Web3}: decentralized platforms whose behavior is recorded as immutable, time-stamped event streams. Despite the richness of this data, the field lacks shared, reproducible benchmarks that capture real-world temporal dynamics, specifically censoring and non-stationarity, across extended horizons. This absence slows methodological progress and limits the transfer of techniques between Web3 and broader Web domains.
In this paper, we present the \textit{FinSurvival Challenge 2025} as a case study in benchmarking \emph{temporal Web3 intelligence}. Using 21.8 million transaction records from the Aave v3 protocol, the challenge operationalized 16 survival prediction tasks to model user behavior transitions.
We detail the benchmark design and the winning solutions, highlighting how domain-aware temporal feature construction significantly outperformed generic modeling approaches. Furthermore, we distill lessons for next-generation temporal benchmarks, arguing that Web3 systems provide a high-fidelity sandbox for studying temporal challenges, such as churn, risk, and evolution that are fundamental to the wider Web.
\end{abstract}



\maketitle

\section{Introduction}
\label{sec:introduction}

Benchmarking temporal web analytics poses challenges that are fundamentally distinct from static prediction or classification tasks. 

First, temporal data is inherently \emph{non-stationary}: user behavior, system incentives, and external conditions evolve over time, often invalidating assumptions of identically distributed samples. In web-scale systems, these shifts may be abrupt (e.g., algorithm updates) or adversarial (e.g., spam waves), and have been repeatedly observed in large-scale Web systems~\cite{leskovec2007graph,kumar2006structure,bar2004sic}.

Second, temporal datasets are subject to \emph{censoring and truncation}. Observations often end before an event of interest occurs, either due to limited observation windows or user disengagement. This is particularly acute in Web platform settings, where user lifecycles are incomplete and system rules change mid-stream, and these challenges are well documented in longitudinal Web and user lifecycle analysis~\cite{kleinbaum1996survival,lancaster1990econometric}.

Third, temporal benchmarks must balance \emph{realism and reproducibility}. While realistic benchmarks require long horizons and rich context, reproducibility demands fixed splits and resistance to leaderboard overfitting.
Prior work on benchmarking and shared tasks has shown that leaderboard-driven evaluation can encourage this type of overfitting and exploitation of temporal artifacts, particularly when evaluation protocols are not fixed or opaque~\cite{codabench}.

Finally, temporal benchmarks often lack a clear notion of \emph{ground truth}. Outcomes such as churn, failure, or disengagement may be ambiguous, delayed, or influenced by unobserved interventions. Designing benchmarks that reflect meaningful temporal outcomes without introducing hindsight bias remains an open challenge.
Similar concerns about ambiguous or delayed ground truth have been raised in studies of user churn, content moderation, and system failure on the Web, where outcomes may depend on unobserved interventions or exogenous events~\cite{harrell1996multivariable}.

These challenges are not unique to any single domain. Rather, they manifest wherever longitudinal user interactions, evolving incentives, and incomplete observations intersect at web-scale. In this sense, Web3 systems provide a particularly revealing case study: they amplify temporal effects while making event histories fully observable.

We present the \textbf{FinSurvival 2025 Challenge}, run in conjunction with The Sixth ACM International Conference on AI in Finance (ICAIF) 2025\footnote{\url{https://icaif25.org}}, to demonstrate how Web3 can serve as a high-fidelity sandbox for benchmarking temporal intelligence. Hosted on Codabench~\cite{codabench}, the challenge centered on \emph{time-to-event} (survival) modeling in Decentralized Finance (DeFi), asking participants to predict the elapsed time between an on-chain user action (e.g., borrowing) and a consequential outcome (e.g., repayment or liquidation) in the presence of censoring and temporal drift.

Beyond introducing a competition or dataset, this paper treats benchmarking itself as a methodological tool for studying temporal intelligence. By fixing task definitions, temporal splits, and evaluation metrics under realistic conditions of censoring and non-stationarity, the FinSurvival Challenge functions as a controlled empirical setting for observing how modeling assumptions about time, risk, and behavior interact with real-world temporal dynamics. This perspective shifts benchmarking from a passive comparison exercise to an active lens for understanding the strengths and limitations of temporal modeling approaches on the evolving Web.
This paper contributes:
\begin{enumerate}
    
    \item \textbf{Cross-Domain Positioning:} We connect DeFi survival modeling to core problems in temporal web analytics (e.g., churn, retention, and risk accumulation), clarifying how Web3 event streams serve as a high-fidelity longitudinal testbed.
    \item \textbf{Benchmark Architecture:} We describe the FinSurvival benchmark suite, which contains 16 time-to-event tasks constructed from 21.8 million Aave v3 transactions, including task definitions, temporal splits, censoring treatment, and evaluation protocol.
    \item \textbf{Analysis of Winning Methods:} We analyze the top competition solutions and show that domain-informed, hierarchical feature engineering substantially outperforms end-to-end deep survival models with minimal feature construction under long-horizon non-stationarity.
    \item \textbf{Design Principles:} We distill actionable guidelines for building next-generation temporal web benchmarks, emphasizing longitudinal realism, explicit censoring, fixed evaluation pipelines, and cross-domain transferability.

\end{enumerate}

\section{Related Work}
\label{sec:related_work}

This section situates the FinSurvival Challenge within three intersecting research trajectories: temporal web analytics, Web3 and blockchain-based behavioral data, and survival modeling under censoring.

While each of these areas has seen substantial independent progress, they have largely evolved in isolation. 
The FinSurvival benchmark and challenge aim to bridge these communities by introducing a shared, real-world temporal testbed that enables systematic evaluation of time-to-event modeling in an open, decentralized web setting.
Our goal in this section is to highlight conceptual gaps and methodological disconnects between these literatures that motivate the need for shared temporal benchmarks, rather than to comprehensively catalog prior work.

\subsection{Temporal Web Analytics}

Understanding how the Web evolves over time has been a central concern of the TempWeb community. Early work characterized the persistence and decay of web content, revealing that links and pages exhibit highly heterogeneous lifespans and rapid turnover~\cite{bar2004sic}. Subsequent studies analyzed the temporal evolution of large-scale link graphs, uncovering densification laws, shrinking diameters, and non-trivial growth patterns~\cite{leskovec2007graph}. Complementary research examined temporal interaction patterns in online communities, highlighting bursty activity, churn, and evolving user roles~\cite{kumar2006structure}.

Despite these advances, benchmarking temporal web phenomena remains difficult. Much of the Web~2.0 ecosystem is governed by proprietary platforms that create information silos and limit user control over data, restricting access to fine-grained, longitudinal interaction histories. Longstanding research in decentralized web architectures has emphasized the need to overcome such silos and return data ownership and control to users~\cite{yeung2023decentralization}. Related work on web content reuse and accountability has further shown that even when data is technically accessible, a lack of provenance, policy awareness, and auditability severely constrains meaningful large-scale analysis~\cite{seneviratne2009policy,seneviratne2010remix,seneviratne2012augmenting}. While web archiving initiatives and longitudinal crawls partially address data availability, they often lack reliable user identifiers, complete interaction histories, or enforceable usage semantics. As a result, many temporal web benchmarks rely on synthetic data or coarse-grained aggregates, limiting their realism and applicability to downstream predictive tasks.

\subsection{Web3 Intelligence}

In contrast to Web 2.0 platforms where interaction logs are siloed behind proprietary APIs and governed by centralized intermediaries, Web3 platforms expose many core behaviors as public, time-stamped event streams recorded on blockchains~\cite{seneviratne2023web}.
Web3 systems, and blockchains in particular, offer a fundamentally different data regime for temporal web analysis. Public ledgers provide an immutable, fully time-stamped record of transactions, enabling native provenance, ordering, and auditability at scale. In contrast to earlier decentralized web proposals focused on social networking and data control~\cite{yeung2023decentralization}, blockchains operationalize decentralization through cryptographic guarantees and economic incentives. This has enabled a growing body of work on blockchain data mining, including temporal transaction graph analysis, anomaly detection, and the identification of illicit or risky behavior~\cite{weber2019anti,christin2013traveling,gridley2022significant}.

Within the financial domain, prior studies have examined the \emph{temporal dynamics} of initial coin offerings (ICOs), including token lifecycles and market cycles over time~\cite{9223270}, alongside cross-chain interoperability challenges that shape how assets and events propagate across heterogeneous ledgers~\cite{kang2022blockchain}, and smart contract-based payment infrastructures on the Web for cross-border payment transfers that log transaction workflows as time-stamped, auditable traces~\cite{mridul2024smart}.
In DeFi, research has focused on protocol dynamics and adversarial behavior. \citet{daian2020flash} revealed the prevalence of miner-extractable value (MEV) and high-frequency arbitrage strategies that exploit temporal ordering and latency, while \citet{qin2021cefi} analyzed liquidation cascades and feedback loops across centralized and decentralized systems. Broader system-level analyses have highlighted the fragility and systemic risk inherent in DeFi ecosystems~\cite{angeris2020improved,werner2022sok}.

Complementary work has explored the social and informational dimensions of Web3. Studies of cryptocurrency-related discourse on social media demonstrate how off-chain signals, such as sentiment and coordinated behavior on ``Crypto Twitter,'' can reflect and even anticipate on-chain events~\cite{kang2024deciphering}. 

Furthermore, recent studies have integrated blockchain with Semantic Web technologies to establish temporal traceability for content and entities, thereby facilitating the longitudinal analysis of Web3 ecosystems.
By grounding on-chain and off-chain artifacts in shared vocabularies and knowledge graphs, these approaches enable richer provenance trails that capture how information evolves, how narratives propagate, and how identities or resources can be linked across decentralized platforms and ledgers~\cite{kazenoff2020semantic,seneviratne2022blockchain}. This line of work treats blockchains as durable temporal backbones that complement the ephemeral nature of web content, enabling retrospective auditing and time-aware attribution of events.

Recent work has also highlighted the substantial data integration and engineering challenges involved in making DeFi transaction data usable for large-scale temporal analytics.
Flynn et al.~\cite{flynn2023enabling} demonstrate how cross-language data integration pipelines are required to unify smart contract data, protocol metadata, and transactional traces in order to support scalable analytics across heterogeneous DeFi systems.
These findings underscore that, while blockchains expose rich temporal event streams, transforming raw on-chain data into analysis-ready representations remains a non-trivial prerequisite for benchmarking and modeling.

Relatedly, blockchain-enabled incentive mechanisms have been proposed to improve the \emph{temporal reliability} of web-scale datasets by encouraging sustained, high-quality contributions to data collection and labeling pipelines~\cite{kader2024enhancing}. Such mechanisms can create verifiable histories of who labeled what and when, which is critical for maintaining benchmark integrity under concept drift and adversarial adaptation.

Despite this rich body of work, most existing Web3 and DeFi datasets frame problems as static classification or short-horizon prediction tasks, such as fraud detection or price forecasting~\cite{shamsi2022chartalist,asiri2025graph,luo2024ai,al2020labeled}. They typically ignore censoring, delayed outcomes, and heterogeneous user lifecycles. Consequently, they fail to capture the time-to-event nature of key DeFi phenomena, including liquidation, protocol exit, and long-term user survival. 

Our FinSurvival dataset addresses this gap by explicitly modeling user trajectories as survival processes over large-scale, real-world DeFi data.
Specifically, FinSurvival models user behavior as a temporal process, explicitly incorporating delayed outcomes, right-censoring, and heterogeneous user lifecycles. This shift aligns Web3 analytics more closely with longitudinal web phenomena such as churn, retention, and risk accumulation, which are poorly captured by snapshot-based formulations.

\subsection{Survival Analysis for Web3}

Survival analysis provides a principled framework for modeling time-to-event data under censoring, making it well suited for domains in which observations are incomplete, outcomes are delayed, and event timing carries semantic meaning. Classical statistical approaches, most notably the Cox Proportional Hazards model~\cite{cox1972regression}, have been widely used in biomedical and reliability studies due to their interpretability and strong theoretical foundations, including applications in cancer prognosis, cardiovascular risk modeling, clinical trial analysis, and system failure modeling~\cite{kalbfleisch2002statistical,kleinbaum1996survival,meeker2021statistical}.

Beyond medicine, survival analysis has a long history in economics, engineering, and the social sciences, where it has been used to model unemployment duration, technology adoption, customer churn, and component reliability~\cite{lancaster1990econometric,kleinbaum1996survival}. These settings share key characteristics with Web3 systems: heterogeneous actors, evolving risk profiles, and partial observability due to censoring or delayed outcomes. In such environments, survival models offer advantages over static classification by explicitly modeling \emph{when} an event occurs rather than merely \emph{if} it occurs.

Extensions such as time-varying covariates and competing risks further broaden the applicability of survival analysis by enabling models to account for evolving system states and multiple, mutually exclusive event types. Time-varying covariate models capture changes in an individual’s state or environment over time~\cite{lin2002modeling}, while competing risks frameworks explicitly model settings in which different outcomes may preclude one another, such as failure versus withdrawal or churn versus retention~\cite{fine1999proportional,bakoyannis2012practical}. These extensions are particularly relevant in Web3, where users may exit protocols due to liquidation, voluntary withdrawal, inactivity, or migration to alternative platforms.

More recently, the scope of machine learning methods tailored towards survival analysis has expanded significantly, extending the range of models beyond classical formulations.
Random Survival Forests~\cite{ishwaran2008random} introduced nonparametric ensemble learning to capture nonlinear effects and complex feature interactions. Neural approaches such as DeepSurv~\cite{katzman2018deepsurv} adapt deep networks to optimize the Cox partial likelihood, enabling scalable learning from high-dimensional data. Transformer-based architectures, including SurvTrace~\cite{wang2022survtrace}, further incorporate attention mechanisms to model long-range temporal dependencies and heterogeneous risk patterns.

Empirical studies of DeFi lending protocols have shown that liquidation and protocol exit exhibit heavy-tailed temporal behavior, reinforcing the suitability of survival-based formulations over binary classification
\cite{green2023characterizing}.

Despite these advances, empirical evaluation of survival models has largely relied on clinical or synthetic datasets, which differ substantially from open, adversarial, and highly non-stationary environments such as Web3 and DeFi. In decentralized systems, agents act strategically, protocols evolve rapidly, and economic incentives introduce feedback loops that challenge classical modeling assumptions. The FinSurvival benchmark and challenge provide a rare opportunity to evaluate both classical and modern survival models on a temporally rich, economically meaningful, and fully observable Web3 dataset, helping to surface their strengths and limitations in real-world temporal web settings.

\section{FinSurvival 2025 Challenge and Benchmark Design}

The \textbf{FinSurvival 2025 Challenge} was an open competition designed to advance time-to-event modeling on large-scale DeFi data. The challenge required participants to develop models predicting the duration between a predefined \emph{index event} (e.g., a \texttt{borrow} action) and a subsequent \emph{outcome event} (e.g., \texttt{repay} or \texttt{liquidate}), accounting for right-censoring and heterogeneous user behavior.

The challenge builds upon a growing body of research that utilizes transaction-level traces to model risk and user decision-making in lending protocols \cite{green2022defi, green2024defi}. Earlier work established a pipeline for converting raw on-chain transactions into survival datasets, reporting systematic relationships between loan characteristics and outcomes. The \textit{FinSurvival} benchmark suite \cite{green2025finsurvival} expanded this by automating the creation of 16 tasks from data from the Ethereum mainnet deployment of Aave v2, providing 128 engineered features per instance. 

While our FinSurvival benchmark paper \cite{green2025finsurvival} introduced the task suite, dataset construction pipeline, and baseline survival models, the work presented in this paper serves a distinct and complementary purpose.
Specifically, in the FinSurvival challenge, using the FinSurvival benchmark code, we created a different dataset with 90 engineered features. Instead of using the Aave v2 Ethereum mainnet deployment like in the original FinSurvival benchmark as described in \cite{green2025finsurvival}, we used the Aave v3 Polygon subgraph as the source for this dataset.
In the FinSurvival challenge, we accompanied the updated dataset with starter code covering several survival modeling approaches, enabling the community to refine these methods and systematically probe the capabilities and limitations of non-linear survival models.

More importantly, this paper does not introduce a new benchmark per se, but instead examines FinSurvival as a \emph{methodological instrument} for studying temporal web intelligence under realistic conditions of censoring, non-stationarity, and strategic behavior.
Specifically, by organizing the challenge, our contribution lies in (i) analyzing how different modeling choices interact with temporal structure and evaluation constraints, (ii) characterizing why certain classes of methods succeed or fail under long-horizon temporal splits, and (iii) distilling design principles for future temporal web benchmarks that extend beyond the DeFi domain.
In this sense, the FinSurvival Challenge is treated not as an end product, but as a controlled empirical lens for understanding broader challenges in temporal web analytics.

\subsection{Task and Evaluation}
The core objective is \emph{time-to-event prediction} under censoring. Given an index event, models predict the time elapsed until an outcome event, evaluated using the \textbf{Concordance Index (C-index)}~\cite{harrell1996multivariable}. 
The C-index measures whether the model's predictions correctly order comparable pairs of instances.
For a pair of users $i$ and $j$ where user $i$ churns (or defaults) earlier than user $j$ ($t_i < t_j$), a concordant set of predictions assigns a higher risk score (e.g., predicted hazard or log-risk) for $i$ than for $j$.

\begin{equation}
    C = \frac{\sum_{i,j} \mathbb{I}(t_i < t_j) \cdot \mathbb{I}(h_i > h_j)}{\text{Total Comparable Pairs}}
\end{equation}

Here, $t_i$ and $t_j$ denote the observed event times for users $i$ and $j$, respectively, $h_i$ and $h_j$ denote the predicted risk scores (e.g., hazard, log-risk, or negative survival time) produced by the model, and $I(\cdot)$ is the indicator function, which evaluates to 1 if its argument is true and 0 otherwise.
Only comparable pairs for which the ordering of event times is known (i.e., not censored) are included in the denominator.

The choice of the $C$-index is motivated by its ability to handle the heavy-tailed distribution of DeFi transactions. 
Unlike Mean Absolute Error, which can be dominated by long-term survivors, the $C$-index focuses on the \textit{relative risk} ordering, making it invariant to the absolute time-scale of market cycles, which is critical given the non-stationarity of the Polygon Aave v3 deployment.
This metric is robust to censoring, as it only compares pairs where the ordering of events is essentially known.
This formulation reflects a general pattern in temporal web analytics where outcomes are sparse, delayed, or censored.

\subsection{The 16 Competition Tasks}
To ensure robustness of the models developed, participants had to submit predictions for all 16 task pairs. The transitions are categorized by their \textbf{starting index event}:
\begin{itemize}
    \item \textbf{Borrow} $\rightarrow$ \{Deposit, Repay, Withdraw, Liquidated\}
    \item \textbf{Deposit} $\rightarrow$ \{Borrow, Repay, Withdraw, Liquidated\}
    \item \textbf{Repay} $\rightarrow$ \{Borrow, Deposit, Withdraw, Liquidated\}
    \item \textbf{Withdraw} $\rightarrow$ \{Borrow, Deposit, Repay, Liquidated\}
\end{itemize}

Each task implicitly defines a competing risks setting: for a given index event, multiple mutually exclusive outcomes are possible, while observations may be right-censored if no outcome occurs within the observation window. Rather than collapsing these dynamics into binary labels, the benchmark preserves temporal uncertainty by evaluating models on their ability to correctly rank event times under censoring. This design reflects real-world Web and Web3 systems, where user lifecycles are often incomplete and outcomes unfold over extended horizons.

The competitors' submissions were ranked based on the average of their concordance index values across these 16 task pairs.

\subsection{Dataset and Features}

As was previously mentioned, the FinSurvival 2025 benchmark used for the competition builds on the pipeline established by \citet{green2025finsurvival}, which comprises over \textbf{21.8 million records} derived from Aave v3 on Polygon. The 90 engineered features provided to participants fall into three categories:
\begin{enumerate}[leftmargin=*]
    \item \textbf{User-Centric:} Historical activity, total deposit/borrow sums, and frequency of interaction.
    \item \textbf{Market-Level:} Total protocol liquidity, average deposit amounts, and asset prices.
    \item \textbf{Transactional:} Transaction size in USD and cyclical temporal encodings (sine/cosine encodings of time-of-day).
\end{enumerate}

Such feature groupings mirror common abstractions in web analytics, separating user state, global context, and interaction-level signals.
The provided feature set intentionally balances expressiveness and accessibility. Features were designed to capture user history, market context, and transactional signals without requiring participants to reconstruct full transaction graphs or protocol internals. This design choice lowers the barrier to entry while still exposing meaningful temporal structure. Importantly, no forward-looking features or post-outcome signals were included, ensuring that all covariates reflect information available strictly prior to the event of interest.

The training data spans from March 12, 2022, to July 2, 2024 and the test data from July 3, 2024, to August 21, 2025. We chose a temporal split to avoid leakage in the test set, as it would be possible for the model to learn market trends from the training data that it would not be able to learn in a real deployment, where it only has access to past data and will be applied to the most up-to-date data.

The full dataset used in this challenge is publicly available on Codabench.\footnote{\url{https://www.codabench.org/datasets/489750/}}

\subsection{Reproducibility and Fairness}
To ensure reproducible and fair evaluation, all participants were evaluated on identical train-test splits using a fixed evaluation pipeline hosted on Codabench. The top-ranked teams were required to submit open-source code and a concise technical report, enabling independent verification of results and methodological transparency. Their final results were verified by the challenge organizers by running and evaluating the submitted open source code.

\section{Comparative Analysis of Competing Modeling Strategies}

The competition saw diverse approaches, with tree-based models emerging as the preferred architecture due to their ability to handle the non-linear dynamics of the DeFi dataset. 
All submitted solutions are publicly available via Codabench.\footnote{
\url{https://www.codabench.org/competitions/10561/\#/results-tab}}
Among the top 16 contestants, the average C-index was 0.750. 

\subsection{Top-Performing Solutions}

The top three solution approaches are summarized in \Cref{tab:solution-comparison} and are elaborated below.

\begin{table}[t]
\centering
\caption{Comparison of Top-Performing Solutions in the FinSurvival 2025 Challenge}
\label{tab:solution-comparison}
\rowcolors{2}{gray!10}{white}
\begin{tabularx}{\columnwidth}{l >{\centering\arraybackslash}X >{\centering\arraybackslash}X >{\centering\arraybackslash}X}
\toprule
\rowcolor{white}
 & \textbf{1st Place} & \textbf{2nd Place} & \textbf{3rd Place} \\
\midrule
Team & Balancehero & AutoFinSurv & FinBoost \\
Focus & Feature Engineering & Optimization & Ensemble \\
Model & XGBoost AFT & XGBoost Cox & XGBoost Mix \\
Features & Hierarchical (10,000+) & Original\quad (90) & Augmented (90+) \\
Average C-index & 0.914 & 0.849 & 0.847\\
\bottomrule
\end{tabularx}
\end{table}

\subsubsection*{First Place: Hierarchical Feature Engineering~\cite{shin2025finsurvival}}
Team \textbf{Balancehero India (Afinit)} achieved the top rank in the competition with an average C-index of 0.914 across the 16 tasks. Their winning solution centered on a \emph{hierarchical feature engineering framework} designed to explicitly reconstruct user and market state over time under strict temporal constraints.
The team rebuilt transaction histories at multiple levels of abstraction, generating over 10,000 features across 12 interrelated tables. These features captured user-level behavioral trajectories, market-level context, and transaction- and liquidation-specific dynamics. A multi-level indexing scheme (e.g., \texttt{user\_index}, \texttt{market\_index}, \texttt{trx\_index}) preserved relational and temporal integrity at scale, enabling efficient joins and aggregation without introducing leakage.

A key differentiator of this approach was its advanced missing-data recovery strategy. By inferring unobserved or partially logged transactions and liquidation events using temporal ordering and count-based validation, the team reconstructed a more complete behavioral history, which proved especially critical for liquidation-related outcomes where missing events can obscure true risk dynamics.

For modeling, the team employed a two-stage XGBoost Accelerated Failure Time (AFT) framework~\cite{barnwal2022survival}. In the first stage, models were trained across shared outcome types to learn general survival patterns, effectively transferring temporal structure across index events. In the second stage, models were fine-tuned for specific index--outcome pairs, allowing specialization to task-specific dynamics while retaining robust global patterns.

Notably, this combination of temporally consistent feature engineering, leakage-aware data reconstruction, and staged survival modeling yielded a substantial performance margin over all other submissions across most tasks, with particularly strong gains on liquidation events. These results underscore that, in long-horizon and censoring-heavy settings, explicitly encoding temporal state and behavioral history can be more impactful than increasing model complexity alone.

\subsubsection*{Second Place: Automated Optimization~\cite{jian2025autofinsurv}}

The second-place winning team introduced a solution called \textbf{AutoFinSurv} that prioritized automation and hyperparameter optimization and achieved an average C-index of 0.849.
The team focused on a single, highly optimized model class: XGBoost with a Cox proportional hazards objective~\cite{cox1972regression}.
Using Optuna~\cite{akiba2019optuna}, they performed Bayesian optimization across all 16 tasks. Their approach proved highly robust, maintaining consistent C-index scores between the development and final leaderboards.
While the Optuna optimization is computationally expensive, it is largely optional as the contestant also provided a set of base hyperparameters that are nearly as performant as the optimized parameters.

\subsubsection*{Third Place: Strategic Ensembling~\cite{nepal2025finboost}}

The third-place winning team introduced a solution called \textbf{FinBoost}, where they conducted a systematic comparison between classical statistical models (such as Weibull AFT)~\cite{kalbfleisch2002statistical} and deep learning survival models (DeepSurv)~\cite{katzman2018deepsurv}. This method averaged a C-index of 0.847, just slightly behind the second place.
While tree-based models performed best, FinBoost successfully implemented an \emph{ensemble strategy}, combining ``Conservative'', ``Balanced'', and ``Aggressive'' risk postures through performance-weighted averaging.
No single temporal assumption fits all user behaviors; ensembling distinct risk postures provides stability.

\subsection{Performance Analysis}

\subsubsection{Feature Engineering Dominated End-to-End Models}

A striking outcome of the FinSurvival Challenge is that approaches centered on extensive, domain-informed feature engineering substantially outperformed end-to-end deep survival models, even those explicitly designed for temporal data.
This result may appear counterintuitive given recent successes of deep learning in sequential modeling; however, it can be explained by several structural properties of the task and data.

First, DeFi user behavior is highly stateful but sparsely observed.
Critical latent variables such as risk tolerance, liquidation aversion, and responsiveness to market volatility are not directly observable, but must be inferred from long-term behavioral aggregates.
Hierarchical feature construction effectively externalizes this inference process by encoding cumulative exposure, rolling statistics, and cross-market dependencies as explicit covariates.

Second, the use of a strict temporal train-test split amplifies the impact of non-stationarity.
End-to-end models trained on earlier market regimes are prone to overfitting historical patterns that do not generalize under shifting protocol dynamics and macroeconomic conditions.
In contrast, feature-engineered representations that normalize behavior relative to contemporaneous market context exhibit greater temporal robustness.

Finally, survival objectives in tree-based models such as AFT and Cox variants of XGBoost naturally align with the concordance-based evaluation metric.
These models directly optimize ranking consistency under censoring, whereas deep architectures often require careful regularization and calibration to avoid instability in risk ordering.

\subsubsection{Temporal Assumptions as the Hidden Variable}

Beyond model architecture, the FinSurvival Challenge reveals that the primary source of performance differentiation lies in how temporal assumptions are encoded.
Each submission implicitly defines a theory of time: what constitutes short-term versus long-term memory, how rapidly risk accumulates, and which historical events remain salient.

Hierarchical feature engineering imposes explicit temporal structure through aggregation windows, decay functions, and state reconstruction.
By contrast, end-to-end neural models must infer these temporal abstractions from raw sequences, which is challenging under sparse, irregularly sampled event streams and heavy censoring.

This observation suggests that benchmark outcomes are shaped not only by model capacity but by the degree to which temporal semantics are made explicit to the learner.
For temporal web benchmarks, this has important implications: without careful specification of temporal representation choices, comparisons between methods may conflate modeling power with implicit assumptions about time.

\section{Discussion}

Taken together, the winning approaches from the FinSurvival Competition suggest that temporal performance gains in Web3 benchmarks arise less from novel model architectures and more from how temporal structure is represented, aggregated, and aligned with evaluation objectives. This observation reinforces the central role of benchmark design in shaping progress in temporal analytics.

\subsection{Principles for Next-Generation Temporal Web Benchmarks}
\label{sec:principles}

Drawing on the FinSurvival experience, we propose the following principles for future temporal web benchmarks:

\begin{enumerate}
    \item \textbf{Longitudinal First:} Benchmarks should prioritize extended temporal horizons over snapshot prediction tasks.
    \item \textbf{Explicit Censoring:} Temporal incompleteness must be modeled explicitly rather than filtered out.
    \item \textbf{Fixed Evaluation Pipelines:} Reproducibility requires centralized, immutable evaluation protocols.
    \item \textbf{Behavioral State Modeling:} Benchmarks should encourage representations that capture evolving user or system states.
    \item \textbf{Cross-Domain Transferability:} Task formulations should be abstract enough to generalize across domains.
    \item \textbf{Open Infrastructure:} Data, code, and evaluation logic should be openly accessible to support cumulative research.
\end{enumerate}

Adhering to these principles can help establish temporal benchmarking as a foundational pillar of temporal web analytics research.

\subsection{Lessons from DeFi Survival Benchmarks}
\label{sec:defi_lessons}

DeFi can serve as a high-fidelity testbed for temporal web analytics. Unlike many web domains where behavioral logs are fragmented across proprietary platforms, DeFi protocols expose publicly verifiable, time-stamped event streams that capture end-to-end interaction histories at scale.

Several lessons emerge. First, \textbf{temporal context dominates static signals}. Winning solutions consistently emphasized longitudinal user histories over instantaneous transaction features, suggesting that behavioral trajectories matter more than isolated actions.

Second, \textbf{feature engineering encodes temporal assumptions}. Hierarchical aggregation windows, rolling statistics, and state reconstruction implicitly define how models interpret time. The strong performance of feature-rich approaches highlights that temporal benchmarks must specify not only tasks, but also assumptions about temporal representation.

Third, \textbf{evaluation stability matters}. The use of fixed train-test splits and a centralized evaluation pipeline prevented leaderboard leakage and enabled fair comparison, addressing a common failure mode in temporal competitions.

Finally, DeFi benchmarks surface \textbf{adversarial and strategic timing}. Users may accelerate or delay actions in response to market conditions, liquidation thresholds, or protocol changes. Temporal benchmarks that ignore such strategic behavior risk oversimplifying real-world dynamics.

\subsection{Cross-Domain Implications for Temporal Web Analytics}
\label{sec:transfer}

While FinSurvival is rooted in DeFi, its benchmarking principles generalize to many temporal web analytics domains. Web archives, online communities, recommendation systems, and decentralized personal data stores all exhibit longitudinal interaction patterns subject to censoring and drift.

Time-to-event abstractions naturally extend to tasks such as content decay, user churn, moderation interventions, topic emergence, and policy compliance lifecycles as shown in \Cref{tab:mapping}. Similarly, survival-based evaluation metrics provide robustness against incomplete observation windows, which are common in web data.

Moreover, the emphasis on reproducible pipelines and open evaluation infrastructure is directly applicable to large-scale web archives and platform datasets, where access asymmetries often impede benchmarking.

By viewing DeFi not as a niche financial application but as a web-native temporal system, FinSurvival demonstrates how domain-specific benchmarks can inform broader methodological advances in temporal web intelligence.

\begin{table}[t]
\centering
\caption{Mapping DeFi Survival Tasks to Web Analytics}
\label{tab:mapping}
\begin{tabularx}{\columnwidth}{lX} 
\hline
\textbf{DeFi Survival Task} & \textbf{Web Domain Equivalent} \\ \hline
Deposit $\to$ Withdraw & Subscription $\to$ Churn (SaaS) \\
Borrow $\to$ Liquidate & Credit Line $\to$ Default (FinTech) \\
Repay $\to$ Borrow & Feature Engagement $\to$ Upsell (E-commerce) \\
Index Event $\to$ Censoring & User Session $\to$ Inactivity/Logout \\ \hline
\end{tabularx}
\end{table}

\subsection{Limitations}

While FinSurvival offers a realistic and large-scale temporal benchmark, several limitations warrant discussion. First, the benchmark focuses on a single protocol family (Aave v3 on Polygon), which may bias models toward specific market mechanics and liquidation rules. Second, although on-chain data is fully observable, off-chain influences such as user sentiment, governance decisions, or macroeconomic news are not captured. Third, evaluation is limited to concordance-based ranking metrics, which emphasize relative risk ordering but do not assess calibration or absolute time prediction accuracy. Addressing these limitations through multi-protocol datasets, richer contextual signals, and complementary evaluation metrics represents an important direction for future benchmark iterations.

\section{Conclusion}

Temporal Web analytics is increasingly central to understanding how large-scale systems evolve, yet progress remains constrained by the lack of shared, longitudinal benchmarks. 
The FinSurvival Challenge 2025 established that domain-aware feature engineering, particularly when capturing user-level historical states, is critical for modeling DeFi risk. While classical linear Cox models remain a baseline, the success of XGBoost-based solutions suggests that capturing non-linear interactions between market volatility and user history is essential for predictive accuracy. 
Beyond benchmarking, these results suggest promising directions for hybrid approaches that combine automated optimization with structured feature construction, as well as for multi-task and cross-chain survival modeling in DeFi.
Future iterations of this challenge will expand to include multi-protocol data and cross-chain survival dynamics.

Beyond DeFi, the lessons distilled here point toward a broader research agenda for temporal web intelligence, one that treats time not as just another variable, but as a first-class analytic dimension. We argue that investing in open, principled temporal benchmarks is essential for advancing web-scale analytics, enabling fair comparison of methods, and fostering cross-domain insight.
We encourage the TempWeb community to treat benchmarking as a form of methodological infrastructure, one that shapes how temporal phenomena are defined, compared, and ultimately understood across the evolving Web.

In closing, the FinSurvival Challenge demonstrates that progress in temporal web analytics depends as much on how time is operationalized as on the choice of learning algorithm. By exposing censoring, non-stationarity, and strategic behavior within a reproducible benchmarking framework, FinSurvival highlights the central role of temporal representation in shaping model performance. We argue that investing in open, principled temporal benchmarks is not merely a matter of evaluation, but a form of methodological infrastructure that enables the Web research community to reason rigorously about time, behavior, and risk across evolving socio-technical systems.

\subsection*{Challenge Resources}
All challenge resources, including the dataset, starter code, evaluation criteria, winning solutions and their technical reports are available at our FinSurvival 2025 challenge website \url{https://finsurvival.github.io}.

\subsection*{Acknowledgements}

We gratefully acknowledge the FinSurvival Challenge 2025 prize sponsors for their generous support: ChaLearn (1st Place, \$1000), Avalanche (2nd Place, \$500), and Rensselaer Polytechnic Institute (RPI) (3rd Place, \$250). We also thank RPI students Corey Curran and Abid Talukder for their valuable assistance in debugging and validating the challenge infrastructure and evaluation setup.

We acknowledge the support from the NSF IUCRC CRAFT center research grants (CRAFT Grant \#22015) for the research that led to the original FinSurvival benchmark. The opinions expressed in this publication and its accompanying code base do not necessarily represent the views of the NSF IUCRC CRAFT.

\balance 
\bibliographystyle{ACM-Reference-Format}
\footnotesize{\bibliography{references}}
\end{document}